\documentclass[runningheads]{llncs}

\usepackage[final,year=2024]{eccv}

\PassOptionsToPackage{table}{xcolor}
\usepackage{color, soul}
\usepackage{colortbl}
\definecolor{mygray}{gray}{.6}
\definecolor{myblue}{RGB}{89,158,254}
\definecolor{mygreen1}{RGB}{81,150,111}
\definecolor{mygreen2}{RGB}{93,174,86}
\definecolor{myred}{RGB}{160,0,0}
\usepackage{xcolor}
\usepackage{multirow}
\usepackage{pifont}
\usepackage{adjustbox}
\newcommand{\ie}{\textit{i.e.}}
\newcommand{\cmark}{\ding{51}}
\newcommand{\xmark}{\ding{55}}
\usepackage{caption}
\usepackage{changepage}
\usepackage{wrapfig}
\usepackage{subcaption}
\renewcommand{\thefootnote}{\fnsymbol{footnote}}

\usepackage{eccvabbrv}

\usepackage{graphicx}
\usepackage{booktabs}

\usepackage[accsupp]{axessibility}  

\usepackage[pagebackref,breaklinks,colorlinks,citecolor=eccvblue]{hyperref}
\usepackage{orcidlink}

\begin{document}

\title{Dyn-Adapter: Towards Disentangled Representation for Efficient Visual Recognition} 

\titlerunning{Dyn-Adapter}

\author{Yurong Zhang\inst{1} \and
Honghao Chen\inst{2} \and
Xinyu Zhang\inst{3} \and Xiangxiang Chu\inst{3} \and Li Song\inst{1,4}$^{\dag}$}

\authorrunning{Y.~Zhang et al.}

\institute{Institute of Image Communication and Network Engineering, Shanghai Jiao Tong University\\
   \and
Chinese Academy of Sciences \\
 \and
Meituan Inc. \\
 \and Cooperative Medianet Innovation Center (CMIC), Shanghai Jiao Tong University } 

\maketitle
\renewcommand{\thefootnote}{}
\footnotetext{The work is done when Yurong Zhang and Honghao Chen were interns at Meituan. $^{\dag}$Corresponding author.}

\begin{abstract}
  Parameter-efficient transfer learning (PETL) is a promising task, aiming to adapt the large-scale pre-trained model to downstream tasks with a relatively modest cost. However, current PETL methods struggle in compressing computational complexity and bear a heavy inference burden due to the complete forward process. This paper presents an efficient visual recognition paradigm, called \textit{Dynamic Adapter (Dyn-Adapter)}, that boosts PETL efficiency by subtly disentangling features in multiple levels. Our approach is simple: first, we devise a dynamic architecture with balanced early heads for multi-level feature extraction, along with adaptive training strategy. Second, we introduce a bidirectional sparsity strategy driven by the pursuit of powerful generalization ability. These qualities enable us to fine-tune efficiently and effectively:  we reduce FLOPs during inference by 50\%,  while maintaining or even yielding higher recognition accuracy. Extensive experiments on diverse datasets and pretrained backbones demonstrate the potential of \textit{Dyn-Adapter} serving as a general efficiency booster for PETL in vision recognition tasks. 
\end{abstract}

\section{Introduction}
\label{sec:intro}

Very recently, large-scale deep neural networks have achieved remarkable advances and attracted growing interest in the vision community~\cite{vit, chu2021Twins, mae, radford2021learning, videomae, scaling, chu2024visionllama}. These colossal models, often with billions of parameters, are pretrained on large datasets (e.g., ImageNet~\cite{imagenet}) and then adapted to a multitude of downstream tasks~\cite{coco, something, hmdb, zhai2019visual, ade20k}, demonstrating unprecedentedly strong capabilities. Such adaptation is usually done via fine-tuning in transfer learning, which typically updates all the parameters of the pre-trained model. However, with the rapidly growing model size, directly fine-tuning these large-scale models can lead to prohibitively expensive storage overhead and computational cost\cite{repadapter,glora}. To rectify this issue,  research endeavours towards reducing the tuning cost using parameter-efficient transfer learning (PETL) methods~\cite{lora, adaptformer, vpt, noah, repadapter}. PETL methods achieve efficient fine-tune by updating only a small number of parameters. By integrating light-weight modules or prepending additional learnable tokens to the input sequence, PETL methods can achieve comparable or even superior performance than full fine-tuning while keeping a significantly reduced parameter cost.

\begin{wrapfigure}{r}{0.60\textwidth}
  \vspace{-12pt}
  \centering
  \includegraphics[width=0.60\textwidth]{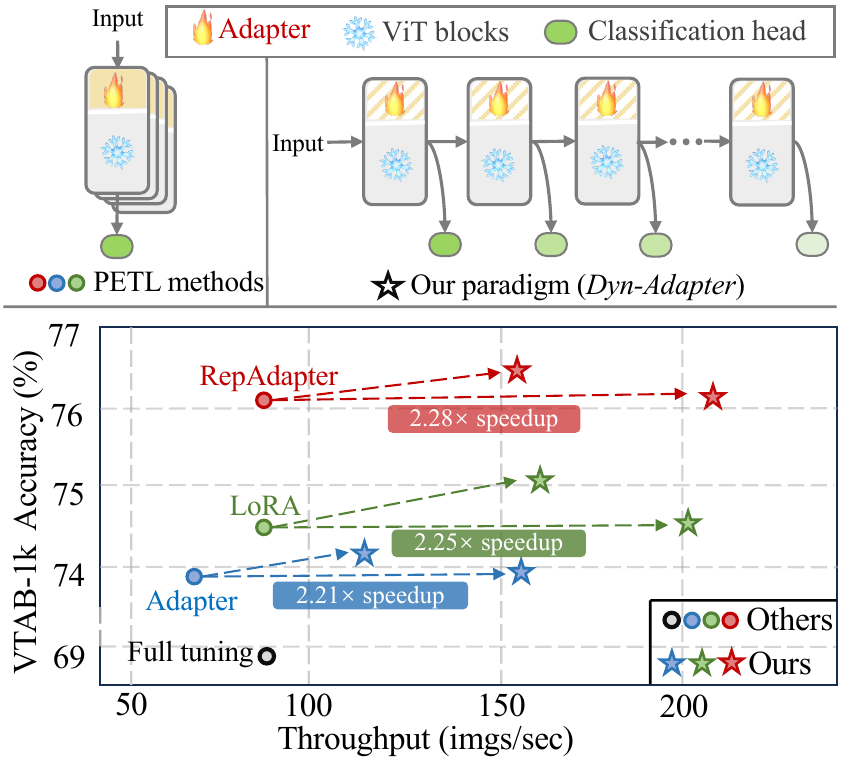}
  \caption{Comparison$_{\!}$ of$_{\!}$ \textit{Dyn-Adapter}$_{\!}$ and$_{\!}$ baselines~(top: para- digm difference, bottom: performance contrast). The throughput is measured on a NVIDIA 3090 GPU with a batch size of 1.}
   \label{poster}
  \vspace{-12pt}
\end{wrapfigure}

Despite the concerned efforts, existing PETL methods suffer from two drawbacks: \textbf{i)} inference efficiency.  Current literature cannot improve the inference efficiency of large-scale models, many methods even introduce additional architecture, resulting in extra latency and FLOPs overhead~\cite{repadapter}.  Therefore, parameter-efficient finetuning can not translate its theoretical advantages into practical efficiency.  Since the application of PETL is usually resource-limited scenarios, this drawback inevitably hinders its development; \textbf{ii)} entangled representation. According to \textit{Information Bottleneck principle (IB)}~\cite{ib1,ib2}, layers close to the input contain more low-level information, while features near the output are rich in semantic meanings.  Although such a learning paradigm achieves great success, it might not be the optimal choice in transfer learning: downstream tasks may suffer from inferior performances if the learned features are over-compressed, or the learned semantic information is irrelevant to the target tasks, especially if there exists a significant domain gap between the source and the target tasks~\cite{revcol}.  For PETL, distinct downstream datasets often possess unique characteristics, such as natural, specialized and structured data, which differ sharply in distribution and composition~\cite{glora}.  According to Dyn-perceiver~\cite{dynamicperceiver}, early classifiers in previous literature force intermediate low-level features to encapsulate high-level semantics and be linearly separable, which essentially means that early classification directly utilizes low-level information while final classifiers focus more on high-level semantics. However, current PETL methods without early exits cannot directly utilize low-level information.

In this paper, we propose a novel PETL framework termed Dynamic Adapter (Dyn-Adpater). Specifically, we propose dynamic-balanced early heads to extract image features from a low level to a high level. These early heads directly act on the intermediate features of different layers and build a connection with the task objective. For samples of different downstream tasks, our approach can dynamically decide which level of features to use depending on the input samples, which can not only improve the accuracy but also reduce unnecessary computation, hence boosting the inference efficiency. Notably, a critical problem in previous early-exiting literature is that early classifiers force intermediate low-level features to encapsulate high-level semantics and be linearly separable, which destroys the inherent low-level feature in shallow layers and invariably backfires the performance~\cite{dynamic4, dynamicperceiver}. In contrast, our approach overcomes this defect fundamentally. Given that the pre-trained backbone parameters are frozen, we speculate that the low-level knowledge in the shallow layers is not interfered, realizing explicit decoupling of low-level feature and high-level semantics. Furthermore, we introduce a bidirectional generalization strategy during the model's forward and backward propagation, which enhances the model's generalization ability and alleviates overfitting. 

Our framework boasts three essential advantages: \textbf{i)} fully explicit decoupling of feature extraction and early classification.  The experiment results in Section 4.2 demonstrate that Dyn-Adapter prominently reduces inference latency with even superior performances; \textbf{ii)} the theoretical efficiency can effectively translate into practical speedup. Remarkably, our framework can eliminate 50\% inference latency and FLOPs of PETL methods without backfiring performance, significantly enhancing their practical efficiency; \textbf{iii)} the simplicity and versatility of our framework. Our approach can be seamlessly migrated into existing PETL methods, consistently outperforming original methods with non-trivial margins.

To evaluate Dyn-Adapter, we apply it to multiple PETL methods including LoRA~\cite{lora}, Adapter~\cite{adapter}, and Rep-Adapter~\cite{repadapter} as Fig~\ref{poster} shows. Extensive experiments across various$_{\!}$ vision$_{\!}$ tasks$_{\!}$ demonstrate$_{\!}$ our method's$_{\!}$ effectiveness.$_{\!}$ For$_{\!}$ instance,$_{\!}$ Our$_{\!}$ designs$_{\!}$ diminish$_{\!}$ RepAdapter's$_{\!}$ 50\%$_{\!}$ inference$_{\!}$ latency and$_{\!}$ FLOPs$_{\!}$ without$_{\!}$ any$_{\!}$ compromise in$_{\!}$ accuracy$_{\!}$ on$_{\!}$ VTAB-1k~\cite{zhai2019visual}.$_{\!}$ Moreover, the visualization results exhibit that our method preserves the low-level features of shallow layers, which further backups our motivation.

\section{Related Work}
\textbf{Parameter-efficient Transfer Learning.} Parameter-efficient Transfer Learning (PETL) aims at fine-tuning a few trainable parameters to transfer large pre-trained models to downstream tasks. PETL was first introduced in the natural language processing (NLP) field~\cite{adapter,lora,nlp2,nlp3,nlp4,nlp5} and extended into large pre-trained vision models across a variety of vision tasks~\cite{cv1,noah,cv3,cv4,adaptformer,ssf,repadapter}. Generally, PETL methods integrate lightweight modules or prepend additional learnable tokens to the input sequence to adapt downstream tasks while keeping the original backbone frozen. For instance, LoRA~\cite{lora} proposes to freeze the pre-trained model weights and inject trainable low-rank decomposition matrices into each layer. VPT~\cite{vpt} proposes to insert a few learnable parameters as prompts and optimize them while freezing the backbone. SSF~\cite{ssf} module scales and shifts features after every MLP, MHSA, and Layernorm module during training, and performs re-parameterization during inference as it is a linear structure. AdaptFormer~\cite{adaptformer} introduces a parallel learnable branch of two linear layers and ReLU over the MLP block and learns only this path while freezing other parts. RepAdapter~\cite{repadapter} inserts sequential lightweight networks into both MHA and MLP, and the additional parameters will be re-parameterized to the nearby projection weights after training.

In this paper, we propose a general framework which is applicable to all existing PETL methods. Without bells and whistles, our Dynamic Adapter can reduce PETL methods' FLOPs up to 50\% without backfiring the fine-tuning accuracy, significantly improving the inference efficiency of PETL methods.

\noindent \textbf{Dynamic Early-exiting For Efficient Visual Recognition.} Dynamic networks~\cite{dynamic1,dynamic2,dynamic3,dynamic4,dynamic5} are designed to improve the inference efficiency of neural networks. Through adapting their computation commensurate with varying input complexities, dynamic networks have demonstrated promising results in efficient visual recognition~\cite{dynamic6}. For instance, \cite{dynamic1} allows examples correctly classified using early layers of the system to exit, and avoid the computational time associated with full evaluation of the network. 
RANet~\cite{dynamic5} proposes a resolution-based dynamic early-exiting framework, which processes simple samples with low-resolution paths and hard samples with high-resolution paths respectively. Despite these advances, a fatal problem exists: classifiers are observed to interfere with each other and significantly degrade the performance of the final exit.  To alleviate this, Dynamic Perceiver~\cite{dynamicperceiver} proposes to decouple early classification and feature extraction with a two-branch structure and a latent code design.  However, since the gradients can still be back-propagated to the shallow layers of the network, these designs can not realize complete decoupling of representations, and the low-level features of the shallow layers are still modified. 

Different from prior literatures, we start from a new perspective, freezing the backbone to keep the feature representation retained.  In our Dynamic Adapter, only the adapter is updated to abstract high-level semantics for classification, while the main backbone is frozen thus its low-level feature can be preserved.  In this way, we realize the fully-decoupling of feature extraction and classification, and the experimental results exhibit the great potential of dynamic early-exiting in the field of transfer learning.

\section{Methodology}

\begin{figure}[t]
    \includegraphics[width=1.0\linewidth]{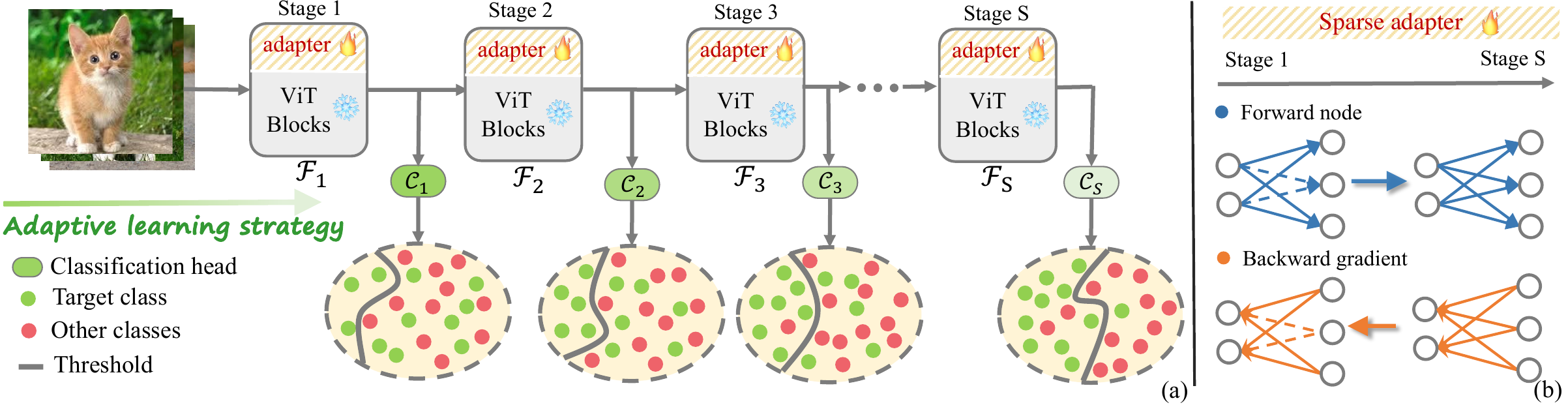}
   \caption{Overview of our \textit{Dyn-Adapter} paradigm. Multiple early supervisions are introduced to facilitate dynamic inference~(section~\ref{secdyn}). Adaptive learning and bidirectional sparsification strategy effectively address \textit{Dyn-Adapter} optimization~(section~\ref{bi}). Dashed lines indicate that the connection between neuron nodes will be dropped during forward or backward.} 
   \label{fig1}
\end{figure}

In this section, we introduce a simple yet elegant paradigm \textit{Dyn-Adapter}. In section ~\ref{review}, we first review the current PETL paradigm. In section~\ref{secdyn}, we introduce the overall framework of \textit{Dyn-Adapter}, including approach setting, early head balance, and adaptive prioritization. Following this, in section~\ref{bi} and \ref{inf}, we present the bidirectional sparsification for more generalized gradient updates and dynamic inference processes. The overall framework is illustrated in Figure~\ref{fig1}.

\subsection{Preliminary}\label{review}
 We first review two representative and top-performing PETL methods, \ie, LoRA~\cite{lora} and RepAdapter~\cite{repadapter}.

\noindent\textbf{LoRA} freezes the pre-trained model weights and leverages trainable low-rank decomposition matrices layer in a parallel way. $\Delta \mathbf{W}$ signifies the learnable low-rank decomposition weights.  Presuming that $\mathbf{W}_0$, $\mathbf{b}_0$, and $\mathbf{X}$ are the pre-trained weights, bias, and input,  respectively, and $g$ denotes a linear layer, then $g(\mathbf{X};\theta) = \mathbf{W}_0\mathbf{X} + \mathbf{b}_0$. The finetuning can be represented as follows and $\Delta \mathbf{W}$ can be reparameterized during inference: 
\begin{equation}
g(\mathbf{X};\theta) = \mathbf{W}_0\mathbf{X} + \Delta \mathbf{W}\mathbf{X} + \mathbf{b}_0 = \mathbf{W}_{\text{LoRA}}\mathbf{X} + \mathbf{b}_0,
\end{equation}
where$_{\!}$ $\theta$$_{\!}$ denotes$_{\!}$ learnable$_{\!}$ weights (\ie, $\mathbf{W}_0$,$_{\!}$ $\Delta\mathbf{W}$$_{\!}$ and$_{\!}$ $\mathbf{b}_0$
),$_{\!}$ and$_{\!}$ $\mathbf{W}_\text{LoRA} = \mathbf{W}_0 + \Delta \mathbf{W}$$_{\!}$ is$_{\!}$ the$_{\!}$ re-parameterized$_{\!}$ weight.

\noindent\textbf{RepAdapter} introduces a sequential adapter to both MHA and MLP, where the upsampling projections are formulated as a group-wise transformation. During inference, the adapter first performs the sequentially structural re-parameterization as follows
\begin{equation}
\begin{split}
f(\mathbf{X};\theta) &= \mathbf{W}_u (\mathbf{W}_d \mathbf{X} + \mathbf{b}_d) + \mathbf{b}_u \\
     &= \mathbf{W}_u \mathbf{W}_d \mathbf{X} + \mathbf{W}_u \mathbf{b}_d + \mathbf{b}_u \\
     &= \mathbf{W}_\text{ada} \mathbf{X} + \mathbf{b}_\text{ada}. \\
\end{split}
\end{equation}
Then, ${f(\mathbf{X};\theta)}$ can be fused into the pre-trained weight $\mathbf{W}_0$ and bias $\mathbf{b}_0$ as
\begin{equation}
\begin{split}
g(\mathbf{X};\theta) &= \mathbf{W}_\text{0}(\mathbf{I} + \mathbf{W}_\text{ada}) \mathbf{X} + \mathbf{W}_\text{0}\mathbf{b}_\text{ada} + \mathbf{b}_0 \\
&= \mathbf{W}_\text{rep} \mathbf{X} + \mathbf{b}_\text{rep}, \\
\end{split}
\end{equation}
where $\mathbf{I}$ is an identity tensor.

\subsection{Dynamic adapter} \label{secdyn}
\textbf{Framework.} Motivated by the demand for high inference efficiency, \textit{Dyn-Adapter} leverages early classification into the PETL methods. The incorporation of early classification allows for dynamic inference depth based on the complexity of the input. The overall framework is illustrated in Fig~\ref{fig1}.
Given an input image $I \in \mathbb{R}^{H \times W \times C}$, Vision Transformer (ViT) preprocesses it into a visual sequence $\mathbf{X}_0\in \mathbb{R}^{ n\times d}$, where $n$ and $d$ denote the token length and embedding dimension, respectively. Then the visual sequence is fed into subsequent $L$ ViT blocks and supervision is performed with an interval of $T$. The total number of introduced supervision $S=L/T$ with $L\mid T$. Specifically, the $i$-th classification head lies after the $l$-th ViT block, where $l=i\cdot T, i\in \{1,2,\cdots,S\}$ and $i$ represents the index of supervision stage. The pre-trained ViT blocks are frozen and only the adapters are updated.

Taking the $l$-th block as example, given the output features of ($l$-$1$)-th block $\mathbf{X}_{l-1}$, the $l$-th ViT block employ computation and the classification prediction $\hat{Y}_l$ is obtained as follows:
\begin{equation}
\begin{aligned}
    \mathbf{X}'_l &= \text{MHA}(g(\text{LN}(\mathbf{X}_{l-1}); \theta)) + \mathbf{X}_{l-1}, \\
    \mathbf{X}_l &= \text{FFN}(g(\text{LN}(\mathbf{X}'_l);\theta)) + \mathbf{X}'_l, \\
    \hat{Y}_l &= \text{HEAD}(\mathbf{X}_l).
\end{aligned}
\end{equation}
where MHA, FFN and LN represent the multi-head attention, feedforward network and layer normalization, respectively.
The objective is to minimize the classification loss between $\hat{Y}_l$ and the corresponding targets $Y$:
\begin{equation}
\begin{split}
    &\mathcal{L} = \sum\nolimits_{i=1}^{S}\lambda_l \: \mathcal{L}_{cls}(\hat{Y}_l, Y),  l = i \cdot T,
\end{split}
\end{equation}
where $\lambda_l$ is the weight of the classification supervision at the $l$-th block, and $\mathcal{L}_{cls}$ is the classification loss.

The core goal of \textit{Dyn-Adapter} is to jointly optimize several early-exit targets in the PETL settings. For simplicity, we denote the ViT blocks from $(i-1)\cdot T + 1$ to $i\cdot T$ as the $i$-th stage, the feature extraction function (including frozen backbone and free adapter) of this stage as $\mathcal{F}_i$, and the classification head of the stage as $\mathcal{C}_i$, where $i \in \{1, 2, \dots, S\}$. The prediction of the ($i$+$1$)-th stage can be represented as $\hat{Y}_{i+1}=\mathcal{C}_{i+1} \circ \mathcal{F}_{i+1}(\mathbf{X}_i)$, where $\circ$ denotes the composition of the functions. The following strategy addresses the \textit{Dyn-Adapter} optimization by comprehensively considering the design of $\mathcal{C}$, $\lambda$, and $\mathcal{F}$. 

\noindent\textbf{Head Balance.} 
Early classification heads $\mathcal{C}$ play a crucial role in dynamic inference while causing gradient interference as commonly acknowledged in supervised learning~\cite{huang2017multi,dynamicperceiver}. When facilitated by the naturally collaborative characteristic of PETL and early supervision, there still exists inconsistency of optimization directions inducted by multiple supervision. 

We study the correlation between different optimization directions and the design of $\mathcal{C}$.   
Since $\mathcal{F}$ and $\mathcal{C}$ are sequentially arranged step by step, we choose the adjacent stages $i$ and $i+1$ for analysis.
The prediction of adjacent stages can be obtained by
$\hat{Y}_i = \mathcal{C}_i(\mathbf{X}_i), \hat{Y}_{i+1} = \mathcal{C}_{i+1} (\mathbf{X}_{i+1}) = \mathcal{C}_{i+1} \circ \mathcal{F}_{i+1}(\mathbf{X}_i)$. 
The outputs of early exit heads share the same optimization target, \ie,
$$ \min\mathcal{L}_{cls}(\hat{Y}_{i}, Y), \min\mathcal{L}_{cls}(\hat{Y}_{i+1}, Y).$$
However, in the past early exit scenarios, the heavy misalignment of the path to obtain $\hat{Y}_i$ and $\hat{Y}_{i+1}$ may cause optimization direction interference intrinsically.

Inspired by above analysis, we propose the ideal requirement for $\mathcal{F}$ and $\mathcal{C}$. Assuming that each part under the classification supervision has been optimized ideally, the relationship between the adjacent $\mathcal{F}$ and $\mathcal{C}$ can be expressed as:
\begin{equation}
\begin{split}
Y & =\mathcal{C}_{i} \circ \mathcal{F}_i(\mathbf{X}_{i-1}) \\
&= \mathcal{C}_{i+1} \circ \mathcal{F}_{i+1}(\mathbf{X}_{i}) \\
&= \mathcal{C}_{i+1} \circ \mathcal{F}_{i+1} \circ \mathcal{F}_i(\mathbf{X}_{i-1}) \\
\end{split}
\end{equation}
which means
$$\mathcal{C}_{i+1} \circ \mathcal{F}_{i+1}(\boldsymbol{\cdot}) = \mathcal{C}_i(\boldsymbol{\cdot}), i \in \{1, 2, ..., S-1\} .$$
Hence, we can conclude that the ideal state of $\mathcal{F}$ and $\mathcal{C}$ should be represented as Eq~\ref{chain} with chain structure
\begin{equation}
    \mathcal{C}_1(\mathbf{\cdot}) \sim \mathcal{C}_2 \circ \mathcal{F}_2(\mathbf{\cdot}) \sim \cdots \sim \mathcal{C}_S \circ \mathcal{F}_S \circ \mathcal{F}_{S-1} \circ \cdots \circ \mathcal{F}_2 (\mathbf{\cdot}).
    \label{chain}
\end{equation}
Note that $\mathcal{F}_i$ is the ViT stage function with considerable complexity, we alleviate the intrinsic interference by leveraging dynamic head in a hierarchical manner. Specifically, we employ classification in the early stages with heavier heads (i.e., MLP layers), while late stages possess lightweight heads for decision. Such design allocates more burden for heads hanging after shallow layers, endowing the network with stronger potential for joint optimization, which is enlightened by the theoretical perspective.


\noindent\textbf{Adaptive Prioritization.} We introduce adaptive weight prioritization for multi-stage learning. Supervision of different stages plays various roles in the joint learning process: \textbf{i)} 
due to the sufficient semantic features borrowed from the pre-trained model in the deep layers, the late classification heads do better in hard sample classification, while shallow layers prefer easier ones. \textbf{ii)} The insertion of early supervision aims to improve inference efficiency, and$_{\!}$ the$_{\!}$ upper$_{\!}$ bound$_{\!}$ of$_{\!}$ recognition$_{\!}$ is$_{\!}$ still$_{\!}$ determined$_{\!}$ by$_{\!}$ deep$_{\!}$ layers.$_{\!}$ \textbf{iii)}$_{\!}$ The$_{\!}$ optimization guided by the late supervision may influence both shallow and deep gradient updates, which implies it plays a more critical function in the general optimization direction.

Therefore, it is necessary to adaptively adjust the prioritization of $\lambda$.  The design of $\lambda$ should adhere to the following guidelines: \textbf{i)} for harder objectives, \ie, late classification, their prioritization needs to be boosted for better learning potential and possible interference avoidance. \textbf{ii)} No matter in what learning period, the weight of layers handling the recognition upper bound should be guaranteed with high priorities. \textbf{iii)} More generalized layers lay the foundation for the subsequent specific task, such as early classification. 

Guided by the policy, we subtly design the prioritization of $\lambda$. At the beginning period, the weight of the deep supervision $\lambda_{deep}$ should be initialized as a relatively large value to ensure the classification performance of the deep layers, and the weight of shallow layers $\lambda_{shallow}$ is set to a relatively small value to protect the learning process of deeper layers. Subsequently, $\lambda_{shallow}$ can be progressively increased, and $\lambda_{deep}$ gradually declines. The upper bound of $\lambda_{shallow}$ equals to the lower bound of $\lambda_{deep}$ to the extent that the hard sample classification ability is preserved, achieving dynamic inference based on data difficulty while ensuring a strong classification ability.

\subsection{Bidirectional sparsification strategy}\label{bi}

Through the design of the overall framework and training strategy, we inspire \textit{Dyn-Adapter} to arrange $\mathcal{C}$ and $\lambda$ according to its intrinsic nature, facilitating optimization of the dynamic paradigm. For the sake of comprehensive design, we further investigate the feature extraction module $\mathcal{F}$ in \textit{Dyn-Adapter}. When large pre-trained models are used for downstream fine-tuning, overfitting easily occurs, hence enhancing the generalization ability is crucial. In \textit{Dyn-Adapter}, the deep blocks are used for deep feature extraction only, while the shallow features are used for early and late classification, then the characteristic of varied layers comes.

The features extracted from shallow layers are reused by multiple supervisions, thus assuming more functions, while features from deep layers are more free to be supervised by deep classification only. Considering this characteristic, the generalization ability of shallow layers should be enhanced. Therefore, we think about both forward and backward propagation processes deeply and employ a bidirectional sparsification strategy to strengthen the generalization power and robustness.

\noindent\textbf{Forward Process.} During the forward process, task collaboration of shallow blocks may cause suboptimal performance. $p_{shallow}$ and $p_{deep}$ denote the dropout probability of shallow layers and deep layers, respectively. We set $p_{shallow}>p_{deep}$. A more drastic dropout $p_{shallow}$ naturally alleviates the issues caused by collaborative effects. By noticeably dropping some nodes, the nodes in the network acquire relatively task-agnostic capability. $p_{deep}$ is set to a normal value, allowing more nodes to focus on learning high-level classifications.  This setting allows for a more flexible and adaptive network and enjoys several charms: \textbf{i)} multi-path forward combination brought by sparsification implies a voting mechanism, which contributes to more robust features. \textbf{ii)} Sparsification encourages the nodes to learn towards objectives independently, eliminating joint adaptability between neuron nodes and enhancing the generalization capability.

\noindent\textbf{Backward Process.} 
During the conventional back-propagation process, the weights $\mathbf{w}$ of all parameters are updat, which results in a relatively fixed paradigm. To endow the gradient updates with a larger combinatorial capacity and relieve overfitting, we employ a masked gradient update strategy.

Specifically, we randomly generate a gradient mask $M$ with a certain mask probability $p_m$. In the backward process, the gradients corresponding to a mask value of 0 are not updated, while those with a mask value of 1 are updated. Given that the mask is randomly generated each time, there will be a diverse combinations for gradient updates, allowing for a more flexible backward path and stronger generalization capabilities, being mathematically expressed with: 
\begin{equation}
    \Delta \mathbf{w}=\alpha \frac{\partial \mathcal{L}(\mathbf{w})}{\partial \mathbf{w}} \odot M,
\end{equation}
where $\alpha$ represents the coefficient of gradient update, and $\odot$ is the dot product operation.

\subsection{Dynamic Inference}\label{inf}
\begin{wrapfigure}{r}{0.53\textwidth}
\vspace{-10pt}
  \centering
  \includegraphics[width=0.53\textwidth]{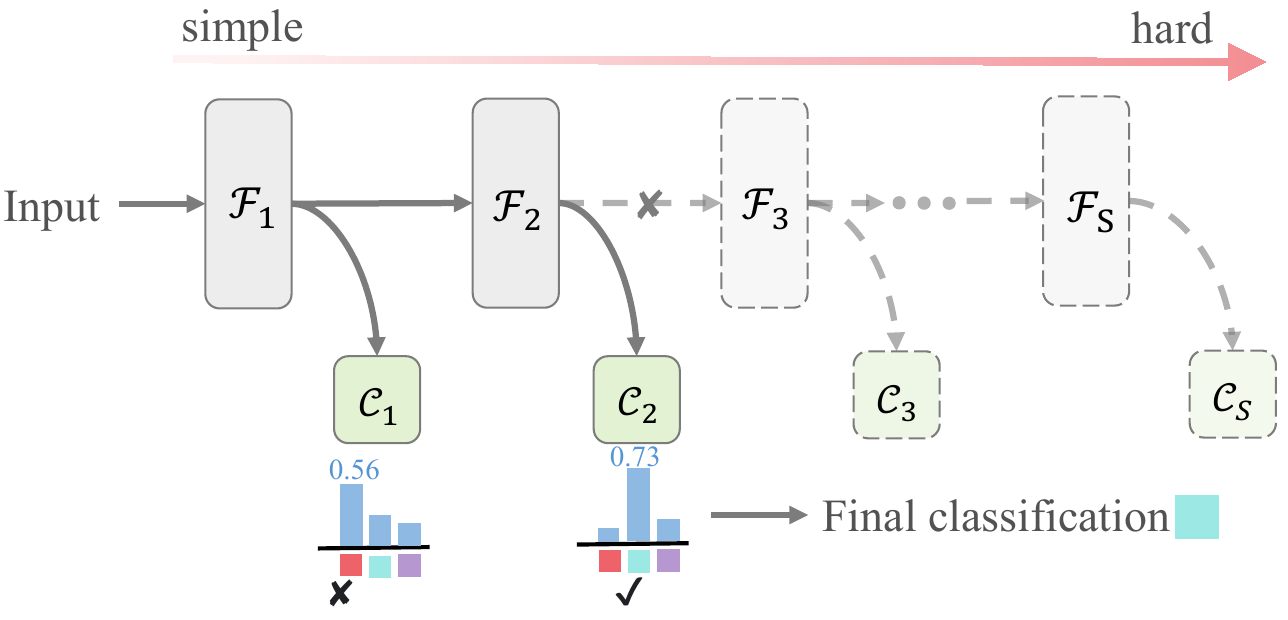}
  \caption{Dynamic inference process.}
  \label{fig:fig2}
  \vspace{-10pt}
\end{wrapfigure}
During inference, we dynamically adjust the network depth based on the complexity level of input samples. As shown in Fig~\ref{fig:fig2}, when the early stage struggles in handling the input sample and the confidence~(the max value of the \textit{softmax} probability) fails to reach the threshold, the network will step into the next stage. Once the confidence exceeds the threshold, the inference process exits. The final classification results are obtained depending on the input characteristic, which is consistent with our design in the training process. 

\section{Experiment}
\label{headings}
\begin{table}[t]
  \caption{Results on VTAB-1k benchmark. ViT-B/16 pre-trained on ImageNet-21k is used as the vision model of all methods. Our framework can reduce PETL methods' FLOPs by 50\% without backfiring performance, or achieve a noticeable improvement with a 30\% reduction in FLOPs. These methods with the prefix Dyn- represent our proposed Dyn-Adapter paradigm employed on different baseline methods. The first row against a gray background demonstrates results with FLOPs constrained to 50\% of baseline. The second row is 70\% of baseline FLOPs.}
  \label{sota}
  \centering
  \resizebox{\textwidth}{!}{
  \setlength{\tabcolsep}{1.5pt}
  \begin{tabular}{lcccc|ccccccc|cccc|cccccccc}
    \toprule
    \multirow{6}{*}{Method} & & & & & \multicolumn{7}{c|}{\textbf{Natural}} & \multicolumn{4}{c|}{\textbf{Specialized}} & \multicolumn{7}{c}{\textbf{Structured}} \\
     & \rotatebox{90}{Param (M)} & \rotatebox{90}{FLOPs (G)}  & \rotatebox{90}{\small{Throughput~(imgs/s)}} &  \rotatebox{90}{Avg. Acc.}
    & \rotatebox{90}{Cifar100} & \rotatebox{90}{Caltech101} & \rotatebox{90}{DTD} & \rotatebox{90}{Flowers102} & \rotatebox{90}{Pets} & \rotatebox{90}{SVHN} & \rotatebox{90}{Sun397} & \rotatebox{90}{Camelyon} & \rotatebox{90}{EuroSAT} & \rotatebox{90}{Resist45} & \rotatebox{90}{Retinopathy} & \rotatebox{90}{Clevr-Count} & \rotatebox{90}{Clevr-Dist} & \rotatebox{90}{DMLab} & \rotatebox{90}{KITTI-Dist} & \rotatebox{90}{dSpr-Loc} & \rotatebox{90}{dSpr-Ori} & \rotatebox{90}{sNORB-Azim} & \rotatebox{90}{sNORB--Ele}  \\
    \midrule
    \textit{Conventional~FT} \\ 
    \hline
    Full tuning~\cite{vpt}  & 85.8 & 16.8 & 91.3 &  68.9  &  68.9 & 87.7 & 64.3 & 97.2  &  86.9  &  87.4   &  38.8 & 79.7 & 95.7 &  84.2 &  73.9 &  56.3  &  58.6 & 41.7 &  65.5 &  57.5 & 46.7 &  25.7 & 29.1 \\
    Linear probe~\cite{vpt} &  0.04 & 16.8 & 90.6  &  57.6 & 64.4  & 85.0 & 63.2 & 97.0  & 86.3   & 36.6 & 51.0 & 78.5 &  87.5 &  68.5 &  74.0  & 34.3 & 30.6 &  33.2 &  55.4 & 12.5 &  20.0  & 9.6  & 19.2 \\ [0.2ex]
    \hline 
    \textit{PETL methods} \\ 
    \hline
    VPT~\cite{vpt} & 0.53 & 22.4  & 87.7 & 72.0 & 78.8 & 90.8 &  65.8 & 98.0 & 88.3  &  78.1 & 49.6 & 81.8 & 96.1 & 83.4 & 68.4  &  68.5  & 60.0 & 46.5 & 72.8 & 73.6 & 47.9 & 32.9  &  37.8 \\
    AdaptFormer~\cite{adaptformer}  & 0.16 & 16.9 & 70.4 &  74.7   &70.8&91.2&70.5&99.1&90.9&86.6&54.8&83.0&95.8&84.4 &76.3&81.9&64.3&49.3&80.3&76.3&45.7&31.7&41.1 \\
    NOAH~\cite{noah} &0.36&16.9& 72.3&75.5&69.6&92.7&70.2&99.1&90.4&86.1&53.7&84.4&95.4&83.9&75.8&82.8&68.9&49.9&81.7&81.8&48.3&32.8&44.2 \\
    SSF~\cite{ssf}&0.24&16.8& 90.3&75.7&69.0&92.6&75.1&99.4&91.8&90.2&52.9&87.4&95.9&87.4&75.5&75.9&62.3&53.3&80.6&77.3&54.9&29.5&37.9\\
    \hline
    Adapter~\cite{adapter} &0.16&16.9& 69.9&73.9&69.2&90.1&68.0&98.8&89.9&82.8&54.3&84.0&94.9&81.9&75.5&80.9&65.3&48.6&78.3&74.8&48.5&29.9&41.6  \\
    \rowcolor[rgb]{0.93,0.93,0.93}
    &\small{+0.12}&\textbf{8.5}& \textbf{154.7}&\textbf{73.9}&68.3&90.8&68.4&98.9&88.7&85.7&54.3&83.5&95.8&84.7&75.8&78.9&64.6&47.4&78.3&74.2&47.1&29.6&39.6 \\
    \rowcolor[rgb]{0.93,0.93,0.93}
    \multirow{-2}{*}{Dyn-Adapter}&\small{+0.12}&\textbf{11.7}& \textbf{116.6}&\textbf{74.2}&68.3&91.1&67.6&98.9&89.5&85.7&54.3&83.0&95.8&84.6&75.8&80.4&64.8&48.5&78.5&76.0&49.4&30.0&40.0 \\
    \hline
    LoRA~\cite{lora} &0.29&16.8& 90.5&74.5&67.1&91.4&69.4&98.8&90.4&85.3&54.0&84.9&95.3&84.4&73.6&82.9&69.2&49.8&78.5&75.7&47.1&31.0&44.0  \\
    \rowcolor[rgb]{0.93,0.93,0.93}
     &\small{+0.17}&\textbf{8.5}& \textbf{204.2}&\textbf{74.5}&67.7&90.5&70.0&99.0&89.4&86.3&53.6&86.2&95.7&84.3&75.0&79.9&67.3&48.5&81.9&77.8&45.4&31.2&38.4 \\
    \rowcolor[rgb]{0.93,0.93,0.93}
     \multirow{-2}{*}{Dyn-LoRA}&\small{+0.17}&\textbf{11.7}& \textbf{159.1}& \textbf{75.0}&67.9&90.5&70.4&99.1&89.8&86.4&53.6&86.3&95.7&84.3&75.1&81.6&67.8&50.2&82.1&79.1&47.0&31.6&39.6  \\
    \hline 
    Repadapter~\cite{repadapter}
&0.22&16.8& 90.9&76.1&72.4&91.6&71.0&99.2&91.4&90.7&55.1&85.3&95.9&84.6&75.9&82.3&68.0&50.4&79.9&80.4&49.2&38.6&41.0  \\
    \rowcolor[rgb]{0.93,0.93,0.93}
    
&\small{+0.16}&\textbf{8.5}& \textbf{207.5}&\textbf{76.1}&71.9&92.2&71.2&99.2&89.9&90.4&54.3&85.7&96.1&86.3&76.1&78.7&68.2&49.8&81.0&82.4&48.5&36.4&41.9 \\
    \rowcolor[rgb]{0.93,0.93,0.93}
    \multirow{-2}{*}{Dyn-Repadapter}
    &\small{+0.16}&\textbf{11.7}& \textbf{156.7}&\textbf{76.4}&71.8&92.6&71.7&99.1&90.6&90.8&54.3&85.8&95.9&86.4&76.1&80.3&68.9&49.9&81.9&82.3&50.3&36.8&41.1 \\
    \bottomrule
  \end{tabular}}
\end{table}

\textbf{Datasets and Metrics.} We leverage VTAB-1k~\cite{zhai2019visual} benchmark to evaluate the transfer learning performance of our approach. VTAB-1k contains 19 dataset subsets, which can be grouped into \textit{Natural}, \textit{Specified} and \textit{Structured} categories. Each subset contains 1000 labeled images, in which 800 images are split into \texttt{train} and 200 images are for \texttt{val}. When inducting a few-shot learning experiment, five fine-grained datasets~(\ie, Food-101, StanfordCars, Flowers102, FGVCAircraft and OxfordPets) are leveraged. For the domain generalization, we train the model on ImageNet and test on four other variants of ImageNet (\ie, ImageNetV2, ImageNet-Sketch, ImageNet-A and ImageNet-R) that perform various
types of domain shift. As for video classification,  We employ Something-Something V2~\cite{ssv2} dataset, a large-scale video classification benchmark with 174 categories, containing 169\textit{k} videos for \texttt{train} and 20\textit{k} videos for \texttt{val}. \textit{Top-1 classification accuracy} is the metric.

\noindent\textbf{Implementation Details.} ViT-Base (ViT/16)~\cite{vit} pre-trained on ImageNet-21k~\cite{imagenet} with supervision is employed as default backbone. We empirically set supervision interval $T=3$, stage number $S=4$ and insert supervisions uniformly. The upper bound of $\lambda_{shallow}$ $\sup(\lambda_{shallow})$ equals to $\inf(\lambda_{deep})$ which is set to 0.5. The hyper-parameter $p_{shallow}$ and $p_{deep}$ are 0.5 and 0.1 respectively, and gradient mask probability $p_m=0.1$. For all models, we trained for 100 epochs. The throughput and GPU latency in this paper are all tested on NVIDIA 3090 GPU. Other details including data augmentation and initialization are consistent with previous work~\cite{lora, repadapter}.

\subsection{Comparison to State-of-the-Arts}
We employ the proposed \textit{Dyn-Adapter} paradigm on three classic baseline methods including Adapter~\cite{adapter}, LoRA~\cite{lora}, and RepAdapter~\cite{repadapter}. As shown in Table~\ref{sota}, our paradigm stably boosts inference efficiency and preserves base accuracy without any compromise. \textit{Dyn-Adapter} maintain or slightly outperform baseline methods in the dramatic 50\% FLOPs decline case. When the inference FLOPs are approximately equal to 70\% of the corresponding baseline, the accuracy is further yielded to a higher level~(\ie, +0.3\% to +0.5\%). Notably, Dyn-RepAdapter set new \textit{state-of-the-art}, surpassing the baseline with 0.3\% accuracy and saving 30\% computational complexity simultaneously, demonstrating the strong adapting ability of \textit{Dyn-Adapter}.  
\subsubsection{Efficiency Analysis}
Inference speed lies in a crucial position in PETL performance analysis. Table~\ref{efficiency} lists FLOPs variation~($\Delta F$) and GPU latency tested on NVIDIA 3090 GPU of several PETL methods. Traditional PETL methods bring an increase in computational complexity due to the inserted module, which causes a latency of varying degrees. LoRA and RepApater smartly design the adapter module and its inserted position to implement a re-parameter strategy during inference, leading to zero FLOPs and latency change. They have gained a significant advantage for this attribute. When it comes to Dyn-RepAdapter, benefiting from dynamic inference based on input, it has achieved a sharp reduction in computational load and latency for the first time, making a new breakthrough in improving reasoning efficiency.

\begin{table}[t]
  \caption{Efficiency comparison of our method and existing PETL methods during inference. Our theoretical efficiency can effectively translate into practical speedup.}
  \label{efficiency}
  \centering
  \small
  \resizebox{1.00\textwidth}{!}{
  \setlength{\tabcolsep}{4pt}
  \begin{tabular}{lc|ccccc}
    \toprule
    \multirow{2}{*}{Method}  &  \multirow{2}{*}{$\Delta$ F~(G)} & \multicolumn{5}{c}{GPU latency (imgs/sec)}
    \\ \cline{3-7}
    &&bs=1 & bs=4 & bs=16 &bs=64 & bs=128  \\
    \midrule
    Full tuning~\cite{vpt}  & 0 & 91.5 & 375.7  & 539.5&568.8&578.3  \\
    \hline
    VPT~\cite{vpt}   & +5.60 & 86.1 \color{myred}{(-5.9\%)} &  283.5 \color{myred}{(-24.5\%)}   & 381.5 \color{myred}{(-29.2\%)}&406.7\color{myred}{(-28.5\%)}&421.6 \color{myred}{(-27.1\%)}  \\
   Adapter~\cite{adapter}  & +0.03 & 70.9 \color{myred}{(-22.5\%)}&306.6 \color{myred}{(-18.3\%)}&504.7 \color{myred}{(-6.4\%)}&533.5 \color{myred}{(-6.2\%)}&552.4 \color{myred}{(-5.8\%)}  \\ 
   AdapterFormer~\cite{adaptformer} &+0.03&71.4 \color{myred}{(-21.9\%)}&309.9 \color{myred}{(-17.5\%)}&508.1 \color{myred}{(-4.2\%)}&546.0 \color{myred}{(-4.0\%)}&555.2 \color{myred}{(-3.9\%)}\\
   NOAH~(500ep)~\cite{noah} &+0.02&72.1 \color{myred}{(-21.2\%)}&312.7 \color{myred}{(-16.7\%)}&492.9 \color{myred}{(-8.6\%)}&523.9 \color{myred}{(-7.9\%)}&534.7 \color{myred}{(-7.5\%)} \\
    \hline 
    Repadapter~\cite{repadapter} &0& 91.5 \color{myblue}{(-0.0\%)} & 375.7 \color{myblue}{(-0.0\%)}&539.5 \color{myblue}{(-0.0\%)}&568.8\color{myblue}{(-0.0\%)}& 578.3 \color{myblue}{(-0.0\%)}\\
    \rowcolor[rgb]{0.93,0.93,0.93}
    Dyn-Repadapter  &-8.30&202.7 \color{mygreen2}{(+121.5\%)}&843.3 \color{mygreen2}{(+124.4\%)}&1228.7 \color{mygreen2}{(+127.7\%)}&1338.9 \color{mygreen2}{(+135.4\%)}&1369.8 \color{mygreen2}{(+136.9\%)}\\
    \bottomrule
  \end{tabular}}
\end{table}
\begin{table}[t]
  \caption{Image classification accuracy for SSL pretrained objectives. Our method is also well suited to contrastive learning (e.g., MoCo-v3) and masked image modeling (e.g., MAE) methods.}
  \label{pretrained}
  \centering
  \small
  \resizebox{\textwidth}{!}{
  \setlength{\tabcolsep}{3pt}
  \begin{tabular}{l|cccccc|cccccc}
    \toprule
    Pretrained objectives & \multicolumn{6}{c|}{MoCo v3} & \multicolumn{6}{c}{MAE} \\
    \hline
    \multirow{2}{*}{Method} &  \multirow{2}{*}{P (M)} & \multirow{2}{*}{F (G)} & \multirow{2}{*}{Acc.} & \multicolumn{3}{c|}{VTAB-1k} & \multirow{2}{*}{P (M)} & \multirow{2}{*}{F (G)} & \multirow{2}{*}{Acc.}  & \multicolumn{3}{c}{VTAB-1k}
    \\ 
    \cline{5-7} \cline{11-13}
    & &&&\textit{Natural} & \textit{Specialized} & \textit{Structured}  &&&&\textit{Natural} & \textit{Specialized} & \textit{Structured} \\
    
    \midrule
    Full tuning~\cite{vpt} &85.8 & 16.8 & 69.55 & 71.95 & 84.72 & 51.98& 85.8   &16.8 & 64.27 & 59.31 & 79.68  & 53.82  \\
    Linear probe~\cite{vpt} &0.04&16.8&59.62&67.46&81.08&30.33&0.04&16.8&32.10&18.87&53.72&23.70 \\
    \hline
    VPT~\cite{vpt} &0.53&22.4&65.23&70.27&83.04&42.38& 0.53   & 22.4 &41.07& 36.02 &60.61&26.57  \\
   Adapter~\cite{adapter}&0.16&16.9&68.18&74.19&82.66&47.69&0.16& 16.9 & 56.36&54.90&75.19&38.98  \\ 
   \hline 
    Lora~\cite{lora}  &0.29&16.8&70.84&69.84&83.71&58.98 &0.29&16.8&70.28&65.99&82.61&62.23 \\
    \rowcolor[rgb]{0.93,0.93,0.93}
     &\footnotesize{+0.17}&\textbf{8.5}&\textbf{72.33}&73.51&85.32&58.16&\footnotesize{+0.17}&\textbf{8.5}&\textbf{68.30}&66.11 &82.94&55.84 \\
    \rowcolor[rgb]{0.93,0.93,0.93}
    \multirow{-2}{*}{Dyn-LoRA} &\footnotesize{+0.17}&\textbf{11.7}&\textbf{73.07}&73.81&85.48&59.92&\footnotesize{+0.17}&\textbf{11.7}&\textbf{70.36}&66.53&84.13 &60.42 \\
    \hline 
    Repadapter~\cite{repadapter}  & 0.22& 16.8 & 72.03 & 71.82&84.27&60.01 &0.22&16.8&69.46& 66.15 & 81.89 & 60.35  \\
    \rowcolor[rgb]{0.93,0.93,0.93}
&\footnotesize{+0.16}&\textbf{8.5}&\textbf{72.11}&73.53&85.57&57.22 &\footnotesize{+0.16}&\textbf{8.5}&\textbf{68.37} &65.54&82.90 &56.67\\
    \rowcolor[rgb]{0.93,0.93,0.93}
    \multirow{-2}{*}{Dyn-Repadapter} &\footnotesize{+0.16}&\textbf{11.7}&\textbf{73.49}&74.69&85.83&59.96&\footnotesize{+0.16}&\textbf{11.7}&\textbf{70.45} &67.05& 83.71&60.60 \\
    \bottomrule
  \end{tabular}}
\end{table}

\begin{table}[t]
  \caption{Results of 16-shot image classification on few-shot learning datasets.}
  \label{fgvc}
  \centering
  \small
  \resizebox{\columnwidth}{!}{
  \setlength{\tabcolsep}{2.5pt}
  \begin{tabular}{lccc|cccccc}
    \toprule
    Method &  Param (M) & FLOPs (G) &  Avg. Acc. 
    & Food-101 & StanfordCars & Flowers102 & FGVCAircraft & OxfordPets \\
    \midrule
    VPT & 0.13 & 22.4  & 72.0 & 72.6 & 56.0  & 99.4&42.5&89.6  \\
   Adapter &0.24&16.9 & 73.2 & 71.7&60.4&99.5&45.2&89.1  \\ 
   LoRA &0.38&16.8 &75.3&72.5&68.2&99.6&47.6&88.7\\
   NOAH (500ep) &6.69&16.9&76.5&76.3&68.6&99.5&49.1&89.0 \\
    \hline 
    Repadapter &0.43&16.8&74.9&74.6&65.7&99.4&44.8&89.8\\
    \rowcolor[rgb]{0.93,0.93,0.93}
    Dyn-Repadapter &\footnotesize{+0.05}&11.6&74.9&73.3&66.6&99.6&45.6&89.3\\
    \bottomrule
  \end{tabular}}
\end{table}

\begin{table*}
\centering
\caption{Ablation studies. In Table~\ref{prior} and ~\ref{drop}, we report the accuracy of different stages(S)/tasks~(\ie, classifications). \textbf{Final} results are obtained by taking a weighted average of the stage accuracy \textit{wrt} exiting probabilities. The proposed method holds better performance on different classification heads, showing stronger generalization ability in different tasks.}
\vspace{-10pt}
\begin{subtable}{.25\textwidth}
\centering
\caption{Componenent.}
\label{comp}
\begin{tabular}{lc}
    \toprule
    Setting &  Acc. \\
    \midrule
    Baseline & 73.6  \\
    + Head Bal. & 74.5~\color{mygreen2}{(+\textbf{0.9})} \\
    + Sparsity.& 75.2~\color{mygreen2}{(+\textbf{0.7})} \\
    \rowcolor[rgb]{0.93,0.93,0.93}
   + Priority &  \textbf{76.4}~\color{mygreen2}{(+\textbf{1.2})}  \\
    \bottomrule
  \end{tabular}
\end{subtable}%
\hfill
\begin{subtable}{.28\textwidth}
\centering
\captionsetup{justification=centering}
\caption{Heavier head position.}
\label{head}
\begin{tabular}{p{1.2cm}p{0.7cm}p{0.7cm}}
    \toprule
    Shallow & Deep &  Acc. \\
    \midrule
    \centering \xmark & \centering \xmark &  75.8  \\
    \rowcolor[rgb]{0.93,0.93,0.93}
  \centering \cmark &\centering \xmark & \textbf{76.4} \\ 
  \centering \xmark & \centering \cmark  & 76.0 \\
   \centering \cmark &\centering \cmark & 76.1  \\
    \bottomrule
  \end{tabular}
\end{subtable}
\hspace{0.1cm}
\begin{subtable}{.35\textwidth}
\centering
\centering
\captionsetup{justification=centering}
\caption{Parameter scaling.}
\label{param}
\begin{tabular}{cccc}
    \toprule
    Method & P(M) &F(G) &  Acc.  \\
    \midrule
&\small{0.29}&16.8& 74.5\\
    \multirow{-2}{*}{LoRA~\cite{lora}}
    &\small{0.52}&16.8& 74.9 \\ 
   \rowcolor[rgb]{0.93,0.93,0.93}
&\small{0.46}&8.5& 74.5\\
    \rowcolor[rgb]{0.93,0.93,0.93}
    \multirow{-2}{*}{Dyn-LoRA}
    &\textbf{0.46}&\textbf{11.7}& \textbf{75.0}  \\
    \bottomrule
  \end{tabular}
\end{subtable}
\vspace{10pt}
\begin{subtable}{.48\textwidth}
\centering
\caption{Priorty.}
\label{prior}
\begin{tabular}{cc|cccc|c}
			\toprule
			 Shallow &Deep & S1 &S2& S3 & S4 & \textbf{Final} \\ \midrule
     \cmark & \xmark & 95.4  & 86.6 & 66.5 & 63.1 & 74.3
    \\
     \rowcolor[rgb]{0.93,0.93,0.93}
			\xmark & \cmark & \textbf{91.2}  & \textbf{82.3} & \textbf{73.8} & \textbf{70.5} & \textbf{76.4}  \\
        \cmark &\cmark &  91.0 & 81.8  & 72.3  & 68.9 & 75.4 \\
   \bottomrule
		\end{tabular}
\end{subtable}
\hfill
\begin{subtable}{.45\textwidth}
\centering
\caption{Dropout rate.}
\label{drop}
\begin{tabular}{c|cccc|c}
			\toprule
			 Drop rate & S1 &S2& S3 & S4  & \textbf{Final} \\ \midrule
    0.3 &  91.4 & 82.4 & 73.0  & 69.9 & 76.1
    \\
    \rowcolor[rgb]{0.93,0.93,0.93}
			0.5 & \textbf{91.2}  & \textbf{82.3} & \textbf{73.8} & \textbf{70.5}  & \textbf{76.4} \\
        0.7 &   90.5 & 81.6 &  72.0 & 68.9 &  75.2 \\
   \bottomrule
		\end{tabular}
\end{subtable}
\end{table*}

\begin{table}
  \caption{Results in domain generalization.}
  \label{domain}
  \centering
  \small
   \begin{adjustbox}{max width=\linewidth}
  \resizebox{0.95\columnwidth}{!}{
  \setlength{\tabcolsep}{2pt}
  \begin{tabular}{lccccccc}
    \toprule
    \multirow{2}{*}{Method} &  \multirow{2}{*}{Param (M)} & \multirow{2}{*}{FLOPs (G)} & \textbf{Source Dataset} 
    &  \multicolumn{4}{c}{\textbf{Target Dataset}}   \\
    &&& ImageNet & -V2 & -Sketch & -A & -R\\
    \midrule
    VPT & 0.82 & 22.4  & 70.5 & 58.0 & 16.4  & 4.6 & 23.2  \\
   Adapter &0.93&16.9 & 70.5 & 59.1 &16.4&5.5 & 22.1 \\ 
   NOAH (500ep) &7.38&16.9&71.7&66.1&24.8&11.9&28.5 \\
    \hline 
    LoRA &1.06&16.8 &70.8&59.3&20.0&6.9&23.3\\
    \rowcolor[rgb]{0.93,0.93,0.93}
    Dyn-LoRA &+2.21&12.4&71.0&59.3&20.7&7.3&22.5\\
    \bottomrule
  \end{tabular}}
   \end{adjustbox}
\end{table}

\subsection{Generalization Experiments}

\textbf{More Pre-trained Objectives.} In Table~\ref{pretrained}, we explore the performance of \textit{Dyn-Adapter} with SSL pretrained objectives, \ie, Moco v3~\cite{mocov3} and MAE~\cite{mae}, which are representative works for contrastive learning and masked image modeling, respectively. The performance on MAE is consistently stable, and the results on Moco v3 are even more outstanding. Under the condition of 50\% FLOPS, our method can boost accuracy by 1.5\% based on LoRA, which signifies a perfect combination of extreme inference speed and notable performance enhancement.

\noindent\textbf{Few-shot Learning.} Following NOAH~\cite{noah}, we conduct 16-shot few-shot learning on five FGVC datasets as Table~\ref{fgvc}. Reducing about 30\% FLOPs, our approach exhibits comparable performance to the baseline under a few-shot condition, demonstrating the robust ability of our method to transfer based on a few samples.

\begin{wraptable}{r}{0.55\textwidth}
\vspace{-30pt}
  \caption{Results of the video classification task. FLOPs are tested on the video with 8 frames.}
  \label{video}
  \centering
  \small
   \begin{adjustbox}{max width=\linewidth}
  \resizebox{0.50\columnwidth}{!}{
  \setlength{\tabcolsep}{2pt}
  \begin{tabular}{lccccccc}
    \toprule
    Method &  Param (M) & FLOPs (G) &  SSv2 \\
    \midrule
    Full tuning~\cite{vpt} &86.04&54.92&53.97\\
    Linear probe~\cite{vpt} &0.07&54.92&29.23\\
    \hline
    VPT~\cite{vpt} & 0.08 & -  & 43.73   \\ 
   AdaptFormer~\cite{adaptformer} &0.15&55.76 &54.70 \\
    \hline 
    RepAdapter~\cite{adapter} &0.15&54.92&55.26\\
    \rowcolor[rgb]{0.93,0.93,0.93}
    Dyn-RepAdapter &\footnotesize{+0.11}&40.05&55.43\\
    \bottomrule
  \end{tabular}}
   \end{adjustbox}
   \vspace{-10pt}
\end{wraptable}
\noindent\textbf{Large-scale Video Classification.} We further evaluate our paradigm on the more complex task - video classification on SSv2~\cite{ssv2} dataset. As shown in Table~\ref{video}, Dyn-Repadapter achieves superior results compared to the baseline, even when operating at merely 70\% FLOPs, demonstrating robust performance in large-scale video classification tasks.

\noindent\textbf{Domain Generalization.} The capacity of out-of-domain generalization becomes the crucial criterion for measuring PETL methods. Fine-tuning on ImageNet with custom 16-shot setting, we evaluate domain generalization ability by directly adapting to four variants of ImageNet with severe domain shift. As implied by Table~\ref{domain}, our approach maintains a comparable performance in the hard domain shift case, stably occupying 75\% computation.
\subsection{Ablation Studies}

We conduct ablation studies based on RepAdapter, controlling FLOPs to about 70\% of the baseline unless noted.

\noindent\textbf{Component Analysis.} We first investigate components of Dyn-Repadapter in Table~\ref{comp}. As shown, the head balance and bidirectional sparsification progressively elevate the baseline to 74.5\% and 75.2\%, revealing the effectiveness of the subtle design. The addition of an adaptive prioritization strategy further boosts the approach to its best performance, yielding an accuracy of 76.4\%.

\noindent\textbf{Head Capacity.} We explore the impact of the weights of shallow and deep classification heads. In Table~\ref{head}, \cmark indicates heavier heads. When the heads hanging on the shallow layers are heavier, the best performance is achieved, which further strengthens our reasoning.

\noindent\textbf{Learning Priority.} Even with detailed theoretical reasoning, we also experimentally prove the necessity of priority setting in Table~\ref{prior}. In the setting where shallow layers and deep layers are both marked with \cmark, they perform cross-optimization art. When the shallow and deep layers are optimized equally, the gradients interfere with each other. When the shallow layers are optimized first, the resulting bias makes it difficult for the deep features to learn, limiting the performance ceiling, and causing sub-optimal accuracy. In contrast, our adaptive priority is the best choice. 

\noindent\textbf{Dynamic Dropout.} For the drop rate of shallow layers, we employ various dropout rates in Table~\ref{drop} and finally find that a large dropout rate~(\ie, 0.5) provides shallow layer more generalization potential.

\noindent\textbf{Parameter Scaling.} Our approach suffers an increased parameter due to the integration of early classification heads. To ensure fairness in parameter count, we amplify the hidden dimension~(8$\rightarrow$16) of LoRA to scale up its parameters in Table~\ref{param}. Despite the classification accuracy enhancement of LoRA with the rise in parameter count, our approach still maintains comparable performance with fewer parameters, even at 70\% FLOPs, which highlights its superiority.

\subsection{Feature Visualization and Analysis}

\textbf{Disentangled Characteristic.} To measure the similarity between the features output by each block and the classification labels, we calculate
the CKA similarity~\cite{cka} for comparison. We visualize the CKA similarity~\cite{cka} of the output of ViT block/intermediate features of Dyn-Perceiver~\cite{dynamicperceiver} from shallow to deep level and label (normalized to $[0, 1]$). 
The first row represents the similarity between our method's labels and features at different levels, while the second row shows the similarity between labels and features at different levels for previous early-exit methods (e.g., Dyn-Perceiver~\cite{dynamicperceiver}). 

\begin{wrapfigure}{r}{0.60\textwidth}
\vspace{-10pt}
  \centering
\includegraphics[width=0.60\textwidth]{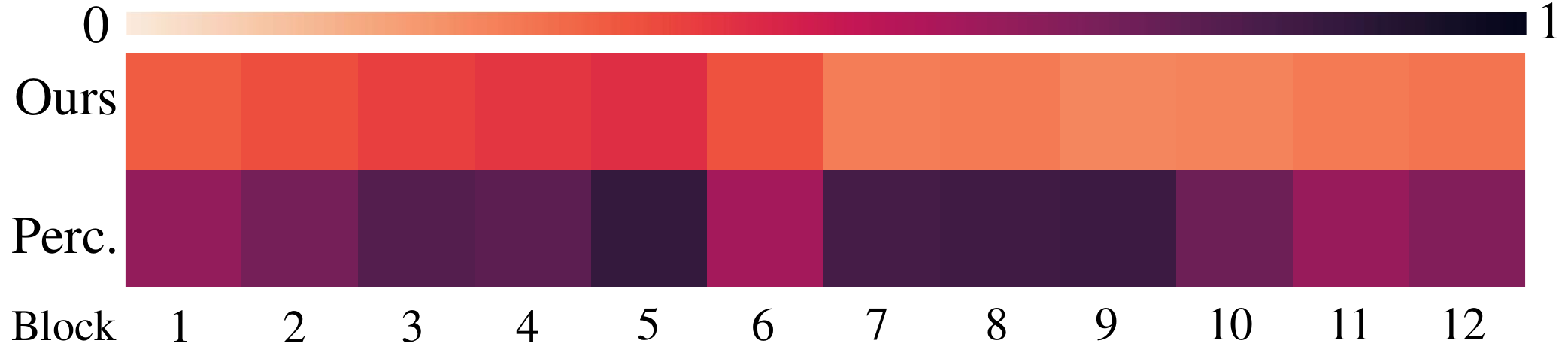}
  \caption{CKA of corresponding features and labels. Perc. is the abbreviation of Dyn-Perceiver.}
  \label{fig6}
  \vspace{-5pt}
\end{wrapfigure}
As shown in Fig~\ref{fig6}, the previous early-exit method~(\ie, Dyn-Perceriver~\cite{dynamicperceiver}) inevitably introduces supervision information into the shallow layers, disrupting the learning paradigm of low-level features in shallow layers and semantic features in higher layers. However, in \textit{Dyn-Adapter}, the adapter module bears the burden of aggregating high-level semantics, while freezing backbone maintains the information from large-scale pre-training, revealing the disentangled characteristic and further facilitating the optimization of \textit{Dyn-Adapter}.

\section{Discussion and Conclusion}
This paper proposes a novel and effective PETL paradigm, \textit{Dyn-Adapter}. We take the leading in exploring PETL with a dynamic inference function, which explicitly decouples feature extraction and early classification and greatly boosts the inference efficiency without accuracy compromise. We subtly design the core component of \textit{Dyn-Adapter} -- early head balance, multi-stage weight prioritization and more generalized feature extraction, comprehensively addressing the adaptive optimization of \textit{Dyn-Adapter}. Our efforts provide a deep insight into promoting inference computation without accuracy decline, which sheds light on the efficient and effective PETL paradigm.

\noindent\textbf{Limitation.} Although we think deeply about the optimization of \textit{Dyn-Adapter} from a theoretical perspective, there lacks profound mathematical modeling for this joint optimization problem to elucidate which direction is more optimal. 

\noindent\textbf{Broader Impact.} This work can reduce the inference efficiency of PETL methods. Under resource-limited circumstances, our method benefits the widespread utilization of PETL and saves hardware resources in practice.

\bibliographystyle{splncs04}

\end{document}